\documentclass[letterpaper, 10 pt, conference]{ieeeconf} 
\usepackage{blindtext, graphicx}
\usepackage{amsmath}
\usepackage{algorithm}
\usepackage[noend]{algpseudocode}
\usepackage{subfig}
\usepackage{graphicx}
\usepackage{caption}
\usepackage[belowskip=-11pt,aboveskip=0pt]{caption}
\usepackage{color}
\definecolor{mygray}{RGB}{200, 200, 200}
\definecolor{mygreen}{RGB}{1, 200, 1}
\definecolor{lightgray}{gray}{0.8}
\definecolor{lightlightgray}{gray}{0.9}

\usepackage[absolute,overlay]{textpos}

\IEEEoverridecommandlockouts                              

\usepackage{cite}
\begin{document}

\begin{textblock*}{20cm}(2.05cm,1.5cm) 
   \color{blue}\Large{\textsf{\textbf{IEEE International Conference on Robotics and Automation (ICRA 2019)}}} \color{black}
\end{textblock*}

\title{\LARGE \bf
Continuous Control for High-Dimensional State Spaces: An Interactive Learning Approach
}

\author{Rodrigo P\'{e}rez-Dattari$^{1}$, Carlos Celemin$^{2}$, Javier Ruiz-del-Solar$^{1}$, and Jens Kober$^{2}$
\thanks{$^{1}$Rodrigo P\'{e}rez-Dattari and Javier Ruiz-del-Solar are with the Electrical Engineering Department and the AMTC,  University of Chile, Santiago, Chile
        {\tt\small (rodrigo.perez.d, jruizd)@ing.uchile.cl}}%
\thanks{$^{2}$Carlos Celemin and Jens Kober are with the Cognitive Robotics Department, TU Delft, Delft, The Netherlands
        {\tt\small (c.e.celeminpaez, j.kober)@tudelft.nl}}%
}

%




\maketitle

\begin{abstract}
Deep Reinforcement Learning (DRL) has become a powerful methodology to solve complex decision-making problems. However, DRL has several limitations when used in real-world problems (e.g., robotics applications). For instance, long training times are required and cannot be accelerated in contrast to simulated environments, and reward functions may be hard to specify/model and/or to compute. Moreover, the transfer of policies learned in a simulator to the real-world has limitations (reality gap). On the other hand, machine learning methods that rely on the transfer of human knowledge to an agent have shown to be time efficient for obtaining well performing policies and do not require a reward function. In this context, we analyze the use of human corrective feedback during task execution to learn policies with high-dimensional state spaces, by using the D-COACH framework, and we propose new variants of this framework. D-COACH is a Deep Learning based extension of COACH (COrrective Advice Communicated by Humans), where humans are able to shape policies through corrective advice. The enhanced version of D-COACH, which is proposed in this paper, largely reduces the time and effort of a human for training a policy. Experimental results validate the efficiency of the D-COACH framework in three different problems (simulated and with real robots), and show that its enhanced version reduces the human training effort considerably, and makes it feasible to learn policies within periods of time in which a DRL agent do not reach any improvement.
\end{abstract}



%
\IEEEpeerreviewmaketitle

\section{Introduction}
In  recent years, outstanding results in complex decision-making problems have been obtained with Deep Reinforcement Learning (DRL). State-of-the-art algorithms have solved problems with large state spaces and discrete action spaces, such as playing Atari games \cite{atari}, or beating the world champion in GO \cite{Silver2016}, along with low level simulated continuous control tasks in environments such as the ones included in the OpenAI Gym \cite{brockman2016openai} and the DeepMind Control Suite \cite{tassa2018deepmind}. Learning policies, parameterized with Convolutional Neural Networks (CNN) for high-dimensional state spaces, such as raw images, gives agents the possibility to build rich state representations of the environment without feature engineering on the side of the designer (which was always necessary in classical RL). These properties can be very useful in robotics, since it is common to find applications with high-dimensional observations, such as RGB images. Giving robots the ability to learn from such high-dimensional data will allow to scale up the complexity of what robots are able to do.

Nevertheless, DRL has several limitations when used to address real world systems \cite{Gu2017}. For instance, DRL algorithms require large amounts of data, which means long training times that in contrast to simulated environments cannot be accelerated with more computational power. If somehow this shortcoming was addressed, sometimes the reward function would still pose a problem as it is hard to specify/model and/or to compute in many cases in the real-world. For instance, sometimes additional perception capabilities to the ones of the agent are needed for computing the reward function, since in theory the reward is given ``by the environment'', not by the agent.

In this regard, the transfer of knowledge learned in a simulator to the real world is a typical solution. However, the mismatch between the virtual and real environment, known as ``Reality Gap'', is often problematic \cite{koos2013transferability}. This results in agents that do not perform at their best in the real world. Thus, it would be preferable to learn/fine-tune policies directly in the real world.

On the other hand, machine learning methods that rely on the transfer of human knowledge to an agent have shown to be time efficient for obtaining good performance policies. Moreover, some methods do not need expert human teachers for training high performance agents \cite{akrour2011preference,Knox:2009:ISA:1597735.1597738,Celemin2018AnInteractive}. This is why they appear to be good candidates to tackle the DRL real-world issues mentioned before. 

Therefore, in this work, we study the use of human corrective feedback during task execution, to learn policies with high-dimensional state spaces, in continuous action problems using CNNs. Our work extends D-COACH \cite{perez2018interactive}, which is a Deep Learning (DL) based extension of the COrrective Advice Communicated by Humans (COACH) framework \cite{Celemin2018AnInteractive}. In the original D-COACH formulation, a demonstration session is required, at the beginning of the training process, for tuning the convolutional layers used for state dimensionality reduction. After that, a fully-connected network policy (connected to the previously trained encoder) is interactively trained during task execution with the human corrective feedback, similarly to the human-agent interaction of the original COACH.

In this paper we introduce an enhanced version of D-COACH, which eliminates the need of demonstration sessions and trains the whole CNN simultaneously, reducing the time and effort of the user/coach for teaching a policy.  In D-COACH no reward functions are needed, and the amount of learning episodes are significantly reduced in comparison to alternative DRL approaches. Enhanced D-COACH is validated in three different problems through simulations and real-world scenarios. In each problem, 
the original and enhanced D-COACH are analyzed and compared with the DDPG method. 

This paper is organized as follows: Section~\ref{sec:related} briefly introduces related Interactive Machine Learning (IML) methods, and Section~\ref{sec:basics} shows the general idea of the original COACH. The enhanced D-COACH is presented in Section~\ref{sec:deepcoach}, followed by the validation experiments in the Section~\ref{sec:result}. Finally, conclusions are drawn in Section~\ref{sec:conclusion}.

\section{Related Work}
\label{sec:related}
IML gathers all techniques in which users interact with agents to shape policies. For instance, Learning from Demonstration (LfD) strategies consist of deriving policies from examples of the state to action mapping (imitation) \cite{Chernova2014,Argall2009} or exploiting the knowledge of a user to teach this mapping from occasional human feedback that could be either corrective or evaluative. Evaluative feedback has been used similarly to RL in methods wherein a human teacher communicates the desirability of the executed action or policy, with validations in problems of state spaces of either low dimensionality \cite{Knox:2009:ISA:1597735.1597738,akrour2011preference,macglashan2017interactive} or high dimensionality \cite{Christiano2017,Warnell2017}. Corrective feedback is given by the teacher directly in the action domain to modify the magnitude computed by the policy. To the best of our knowledge, corrective feedback has been only validated in problems with state spaces of low dimensionality \cite{Celemin2018AnInteractive,Argall2008}, and D\nobreakdash-COACH \cite{perez2018interactive} is the first approach targeting high-dimensional state spaces.

Deep RL showed that it is possible to approach problems with state spaces of high dimensionality \cite{Mnih2015}. In LfD some work has already been done \cite{zhang2017deep}, as well as with evaluative feedback approaches \cite{Christiano2017,Warnell2017}. So, this paper complements these studies in the direction of corrective advice.

Imitation LfD approaches based on data aggregation, such as DAgger \cite{ross2011reduction}, are conceptually close to COACH in the sense that in both cases labels are generated online and as consequence of the states visited by the agent. The main difference is that in the former case the labels are optimal actions, whereas with COACH, corrections are relative improvements over the current agent's actions.


\section{Basics of the Learning Framework}\label{sec:basics}
When teaching with human corrective advice, if the agent executes an action $a$ that the human considers to be erroneous, then s/he will indicate the direction in which the action should be corrected (thus, COACH was proposed for problems with continuous actions). Each dimension of the action has a corresponding correction signal $h$ with values $0$, $-1$ or $1$ which produces an error signal with arbitrary magnitude $e$ that is used to shape directly the policy. Thus, the error is: 
\begin{equation}\label{eq:error}
    \mathrm{error}=h \cdot e
\end{equation}
where $h=0$ indicates that no correction has been advised. $h=\pm 1$ indicates the direction of the advised correction. Before introducing the Deep COACH framework, we describe the algorithm of the original COACH framework that inspired this work.

\subsection*{COACH}
In this framework no value function is modeled, since no reward/cost is used in the learning process \cite{Celemin2018AnInteractive}. A parametrized policy is directly learned in the parameter space, as in Policy Search (PS) RL. 
The classic COACH algorithm shapes two functions parametrized as a linear combinations of basis functions (LCBFs). The objective of the first function is to learn the policy of the agent $\pi(s)=f^{\top}\theta$; the objective of the second function, $H(s)=f^{\top}\psi$, is to learn a prediction of the human feedback. The vector of basis functions $f(s)$, for simplicity is called $f$. The parameter vectors $\theta$ and $\psi$ are updated to shape the models. As it can be seen, $f$ is the same vector for both the Policy Model $\pi(s)$ and the Human Feedback Model  $H(s)$. The Human Feedback Model is used to adapt the size of the error signal that is then used to update the weights of $\pi(s)$. Both functions are updated using stochastic gradient descent each time feedback is received. The pseudocode of COACH is shown in Algorithm \ref{algorithm:COACH}.

\begin{algorithm}[t]
\caption{Basic Structure of COACH}\label{algorithm:COACH}
\begin{algorithmic}[1]
\State \textbf{Require:} error magnitude $e$, human model learning rate $\beta$, time steps $N$
\For{t = 1,2,...,N}{}
\State \textbf{observe} state $s_{t}$
\State \textbf{execute} action $a_{t}=\pi(s_{t})$
\State \textbf{feedback} human corrective advice $h_{t}$
\If{$h_{t}$ is not 0}
\State \textbf{update} $H(s_{t})$ with $\Delta \psi = \beta\cdot (h_{t}-H(s_{t}))\cdot f$
\State $\alpha_{t} = |H(s_{t+1})|$
\State $\mathrm{error}_{t} = h_{t}\cdot e$
\State \textbf{update} $\pi(s_{t})$ with $\Delta \theta = \alpha_{t} \cdot \mathrm{error}_{t} \cdot f$
\EndIf
\EndFor
\end{algorithmic}
\end{algorithm}

\section{COACH for high dimensional state spaces} \label{sec:deepcoach}
The Deep COACH (D-COACH) framework \cite{perez2018interactive} takes the ideas introduced by COACH but, instead of parametrizing policies with LCBFs, it uses deep neural networks for problems of low or high dimensional state spaces. Different architectures may be used depending on the characteristics of the problems. 

In D-COACH, the Human Feedback Model is replaced by a memory buffer from which the agent samples past corrections and replays them, in a similar manner as used in DRL problems \cite{atari}. Both the memory buffer and the Human Feedback Model use information given by past corrections to modify the effects of newer ones. Every time feedback is received, to ensure that recent corrections have an immediate effect on the policy, the network gets updated by that feedback signal, and subsequently the network is updated with a batch sampled from the memory buffer. Also, the network gets updated from the buffer with a fixed frequency every $b$ number of time steps.

Training CNN policies with human feedback is a challenging problem: both the state representation from raw data (this work focuses on applications that observe raw images), and the policy must be learned from data generated by a human teacher. Deep neural networks have shown to need massive sets of data and experience to adequately converge.  The requirement of massive amounts of trials can be problematic because human users cannot keep on assisting the training for a long time due to stress, fatigue, or other factors that may lead to a lack of concentration.  

To tackle this shortcoming, state representation learning strategies are employed. The state representation of the policy is trained with additional criteria, adding an autoencoder to help with the training of the convolutional layers (or encoding layers) of the network. 

\subsection{Basic Deep COACH} \label{sub:P-D-COACH}

\begin{figure}
\centering
\subfloat[][Autoencoder network for learning the state representation of the policy based on the demonstrations database.]{\includegraphics[width=0.48\linewidth]{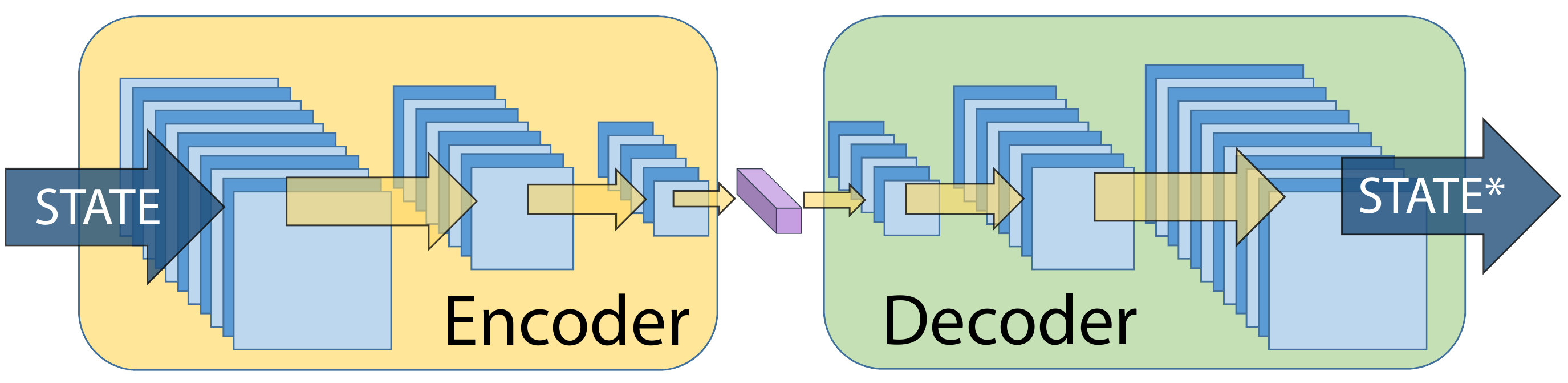}}
\hspace{0.1cm}
\subfloat[][Interactive training of the policy using the trained convolutional layers.]{\includegraphics[width=0.48\linewidth]{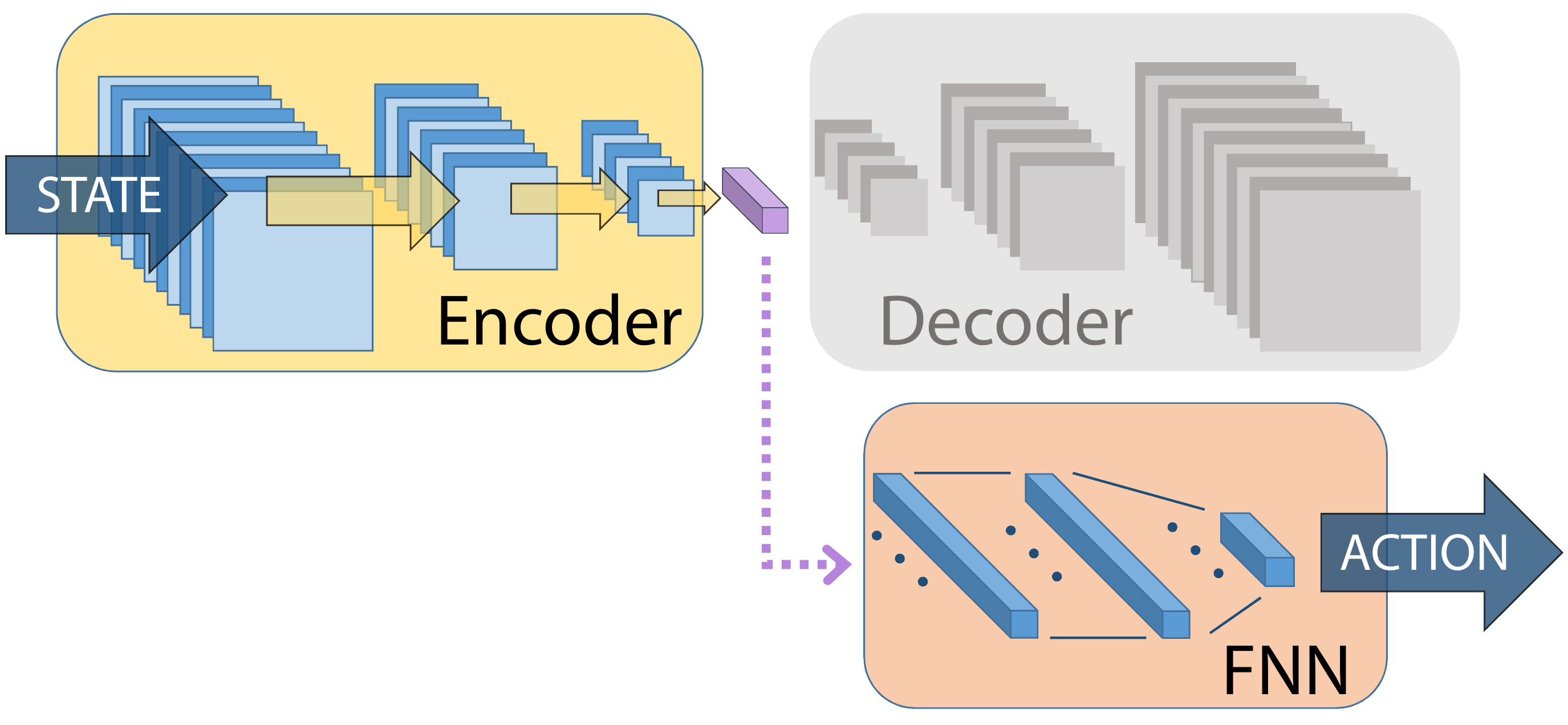}}
\caption{Basic D-COACH.} 
\label{fig:ms} 
\end{figure}

The early version of Deep COACH proposed in \cite{perez2018interactive} is a 3-step sequential process. In the first step a session of demonstrations is recorded. Then in a second step, the convolutional layers are trained  with the recorded data, while tuning an  autoencoder  used  for state representation of reduced dimensionality, (Fig.~\ref{fig:ms}(a)). The state is embedded in the latent space of an autoencoder. Finally, in the third step, the convolutional layers of the trained encoder are frozen, then the subsequent non-convolutional layers (right hand side of Fig.~\ref{fig:ms}(b)) are trained interactively based on the corrections that the teacher advises to the agent. Fig.~\ref{fig:ms} depicts the second and third steps of this strategy. This kind of sequential strategy has been proven to work in decision-making problems \cite{Warnell2017,Finn2015,Ha2018}, but it has the shortcoming that it needs a database with images of the environment to work. This can be time consuming and not robust to changes in the environment.
    
\subsection{Enhanced Deep COACH} \label{sub:E-D-COACH}
In order to eliminate the requirement of recording demonstrations and pre-training an autoencoder, an enhanced version of Deep COACH is proposed, which learns everything in a single interaction step, as the original COACH does. Hence, it allows to train all the parameters of the network interactively from scratch. State representation strategies have been included in order to make the networks converge faster. The basic idea is to train the state representation of the policy with additional auxiliary criteria (autoencoding). So, in addition to the loss function for predicting the policy based on the data generated by the human corrections, it also includes the loss function of the reconstruction at the output of the autoencoder, based on the same data stored during the human corrections. 

\begin{figure}
    \centering
    \includegraphics[width=0.5\linewidth]{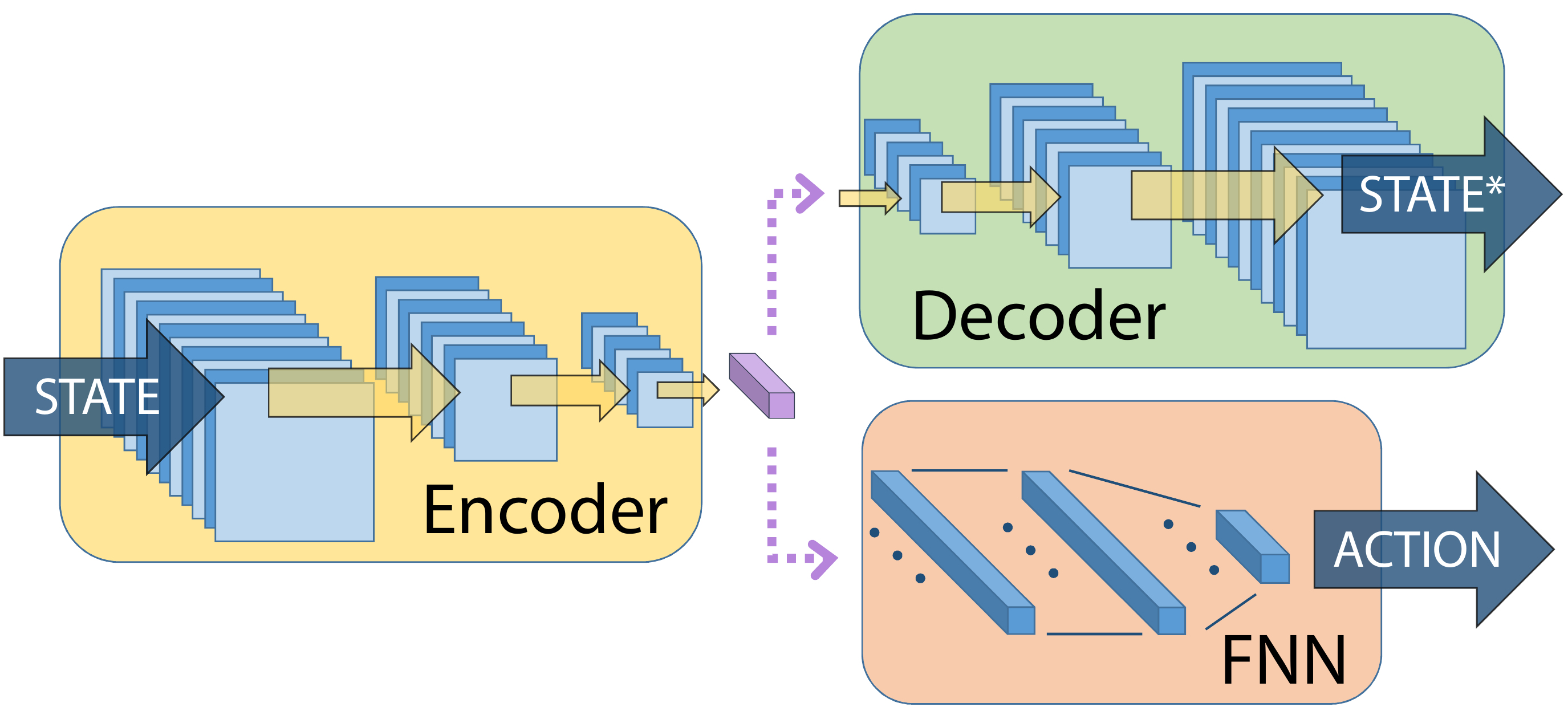}
    \caption{Enhanced D-COACH. The state representation is shared between the autoencoder and the policy training.}
    \label{fig:msim}
\end{figure}

Both networks, the policy and the autoencoder, share the convolutional layers of the encoder as shown in Fig.~\ref{fig:msim}. The policy network has the convolutional layers at the input, followed by a second part that is a fully-connected layer, then this network maps from STATE to ACTION, while the autoencoder network involves the computation from STATE to STATE$^{*}$, wherein  STATE$^{*}$ is the reconstructed image at the output of the decoder, according to Fig.~\ref{fig:msim}.

Algorithm \ref{algorithm:DeepCOACH} describes the enhanced version of D-COACH. The algorithm first sets the hyper-parameters like the magnitude of the error \eqref{eq:error}, and the ones used for the corrections replay. In cases when the teacher advises a correction (line 15), the subsequent lines are evaluated, wherein the policy network is updated along with the autoencoder network.

Two different updates are computed, one for the layers involved in the policy computation, and another for the layers of the autoencoder. When an advice of correction is given, the \textbf{update} \emph{policy} instruction updates the policy network for the current state and the \emph{batch\_update} subroutine is called. This subroutine updates the policy and the autoencoder using a mini-batch sampled from the replay buffer. If the reconstruction error of the autoencoder (difference between STATE and STATE$^{*}$) is greater than a threshold $\epsilon$ (line 6), the autoencoder is updated with the same mini-batch using the instruction \textbf{update} $AE$ (autoencoder). Otherwise, the convolutional layers are frozen, so that the instruction \textbf{update} \emph{policy} in lines 5 and 19, only modifies the non-convolutional layers with the Stochastic Gradient Descent (SGD) operation. The \emph{batch\_update} subroutine is also called every $b$ time steps (line 24).

 The \textbf{update} \emph{policy} instruction uses [state,$y_\mathrm{label}$] pairs (where $y_\mathrm{label(t)}=a_{t}+\mathrm{error_{t}}$), whereas \textbf{update} $AE$ only uses states. The aforementioned condition in line 6 is used for avoiding conflicts in the gradients of both cost functions, so when the latent vector of the AE is considered a good smaller representation of the state, the gradient of the policy must be prevented from harming the learned encoding. Hence, the encoder is kept frozen, unless unknown regions of the state space are visited. 

\begin{algorithm}[t]
\caption{D-COACH }\label{algorithm:DeepCOACH}
\begin{algorithmic}[1]
\algdef{SE}[SUBALG]{Indent}{EndIndent}{}{\algorithmicend}
\algtext*{Indent}
\algtext*{EndIndent}

\State \textbf{Require:} error magnitude $\textit{e}$, buffer update interval $b$, buffer sampling size $N$, buffer max. size $K$, buffer min. size $k$, pre-trained encoder parameters (if 3-step sequential learning) 
\State \textbf{Init:} $B = []$ \emph{\# initialize memory buffer}
\colorbox{lightgray}
{\parbox{\linewidth}{
\State \textbf{function} batch\_update \emph{\# define batch update function}
\Indent
\If{$\mathrm{length}(B) \geq k$}
\State \textbf{update} \emph{policy} using SGD with a mini-batch sampled from $B$
\If{$AE_{\mathrm{error}}>\epsilon$}
\State \textbf{unfreeze} convolutional layers
\State \textbf{update} $AE$ using SGD with a mini-batch sampled from $B$
\Else
\State \textbf{freeze} convolutional layers
\EndIf
\EndIf
}}
\colorbox{lightlightgray}{\parbox{\linewidth}{
\For{t = 1,2,...}{} \emph{\# main loop}
\State \textbf{observe} state $s_{t}$
\State \textbf{execute} action $a_{t}=\pi(s_{t})$
\State \textbf{feedback} human corrective advice $h_{t}$
\If{$h_{t}$ is not \textbf{0}}
\State $\mathrm{error}_{t} = h_{t}\cdot e$
\State $y_{\mathrm{label}(t)} = a_{t} + \mathrm{error}_{t}$ 
\State \textbf{update} \emph{policy} using SGD with pair ($s_{t}$, $y_{\mathrm{label}(t)}$) 
\State \textbf{call} batch\_update
\State \textbf{append} $(s_{t}, y_{\mathrm{label}(t)})$ to $B$
\EndIf
\If{length($B$) $> K$ }
\State $B = B[2:K+1]$
\EndIf
\If{$\operatorname{mod}(t, b)$ is 0}
\State \textbf{call} batch\_update
\EndIf
\EndFor}}
\end{algorithmic}
\end{algorithm}

\section{Experimental Results}
\label{sec:result}
Three different types of experiments were carried out for validating the enhanced D-COACH: i) experiments with simulated teachers for evaluating the learning method under controlled conditions without influence of human factors, ii) validations with real human teachers, and iii) extra validations on real physical systems.

\begin{figure}[h]
    \centering
    \includegraphics[scale=0.35]{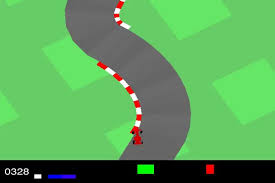}
    \caption{Car Racing, environment view.}
    \label{fig:Car_Racing}
\end{figure}

In these experiments we denote the enhanced D-COACH simply with `D-COACH', while from now on we call the previous version `basic D-COACH'. In the experiments, the learning processes are analyzed in three different problems:

\textbf{Car Racing:} A simulated problem (from OpenAI gym \cite{brockman2016openai}) in which the agent has to learn to drive from a top-down view of a racing car game (see Fig.~\ref{fig:Car_Racing}). The objective of the task is to drive a racetrack as fast as possible without leaving it. The default state that is given by the environment is a $96\times96\times3$ top-down view of the car which we downsampled to $64\times64\times1$. The continuous action space consists of 3 dimensions: \textbf{[direction, acceleration, brake]}. The \emph{direction} range goes from $-1$ to $1$, the \emph{acceleration} from $0$ to $1$ and the \emph{brake} from $0$ to $1$. In this problem, experiments with the simulated teacher and human teachers were carried out. The coupled feedback strategy was used in the experiments with human teachers, as presented in \cite{perez2018interactive}.

\begin{figure}[b]
\subfloat[][Duckiebot.]{\includegraphics[height=0.205\linewidth]{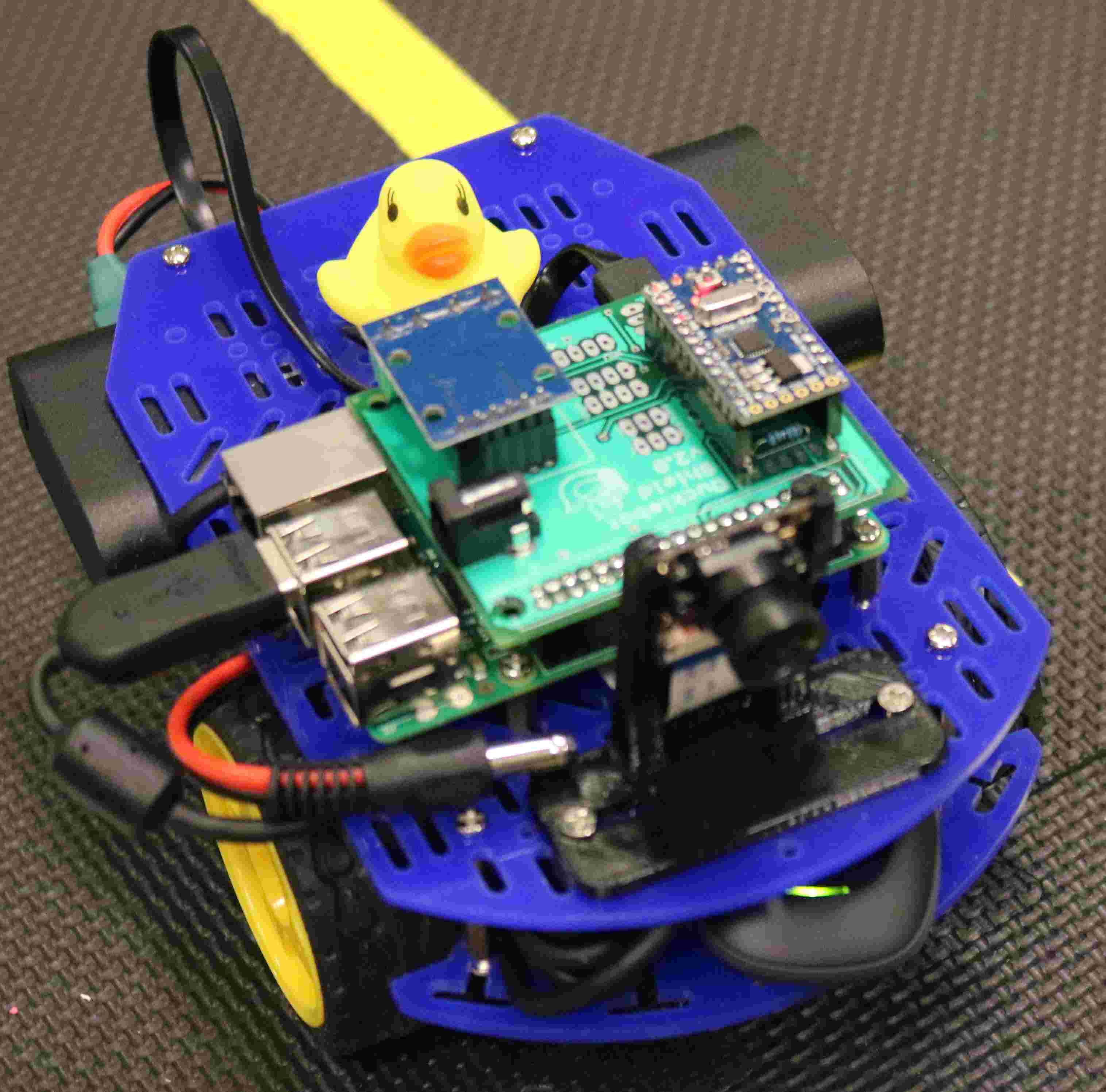}} 
\hfill
\subfloat[][First-person view.]{\includegraphics[height=0.205\linewidth]{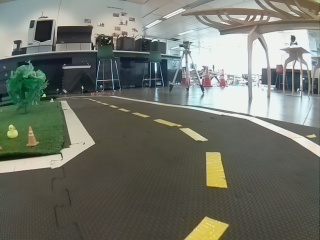}}
\hfill
\subfloat[][First-person simulated view.]{\includegraphics[height=0.205\linewidth]{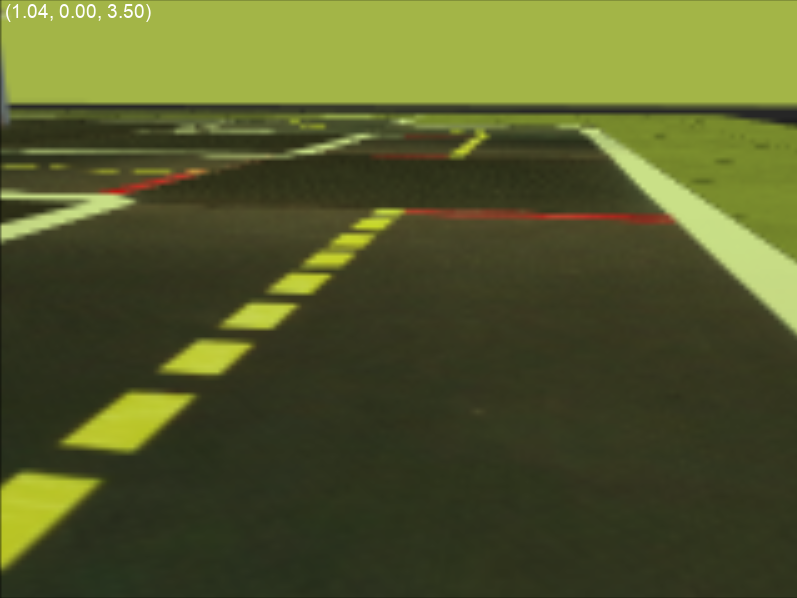}}
\hfill
\subfloat[][Map.]{\includegraphics[height=0.205\linewidth]{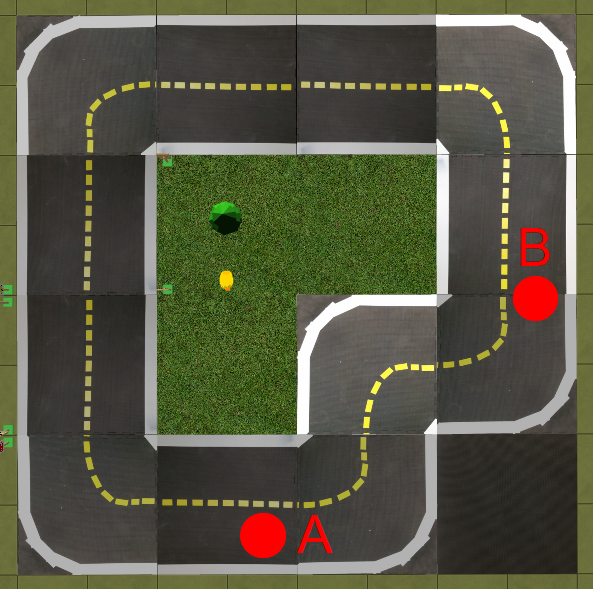}}
\caption{Duckietown.} 
\label{fig:duckietown} 
\end{figure}

\textbf{Duckie Racing:} This is also a driving task, but in this case with a real/simulated robot which has an onboard camera that gives a first-person view of the environment. The real robot consist of a Duckiebot from the project Duckietown \cite{Paull2017}. The simulated robot is based on \cite{gym_duckietown}. A $120\times160\times3$ observation image is received from the environment which is downsampled to $64\times64\times1$. The same road was used for both the simulated and real robot. In the simulations, at the start of each episode, the robot can start, randomly, at the points A or B (plus random noise) of the map (see Fig.~\ref{fig:duckietown}). Each simulated episode lasts 1000 time steps (unless the robot leaves the road before) and as a performance metric a modified version of the default reward function of the environment is used, which has the following shape: $R = Cv\theta - Dd$. $C$ and $D$ are constants ($C=100$, $D=1$), $v$ is the linear velocity of the duckiebot, $\theta$ is its orientation with respect to $\gamma$ (a bezier curve that defines the path the agent is expected to follow) and $d$ is its distance to $\gamma$. The duckiebot is a differential robot, so the default actions consisted of speed commands ranging from -1 to 1 for each of the two wheels. To make it more intuitive for a human teacher to give feedback, the environment had an inverse kinematics module for the actions to be linear and rotational speeds instead, also ranging from -1 to 1. This problem is also used for experiments and validation with simulated and real human teachers.

\begin{figure}[t]
\null\hfill
\subfloat[][3DoF robot arm.]{\includegraphics[height=0.28\linewidth]{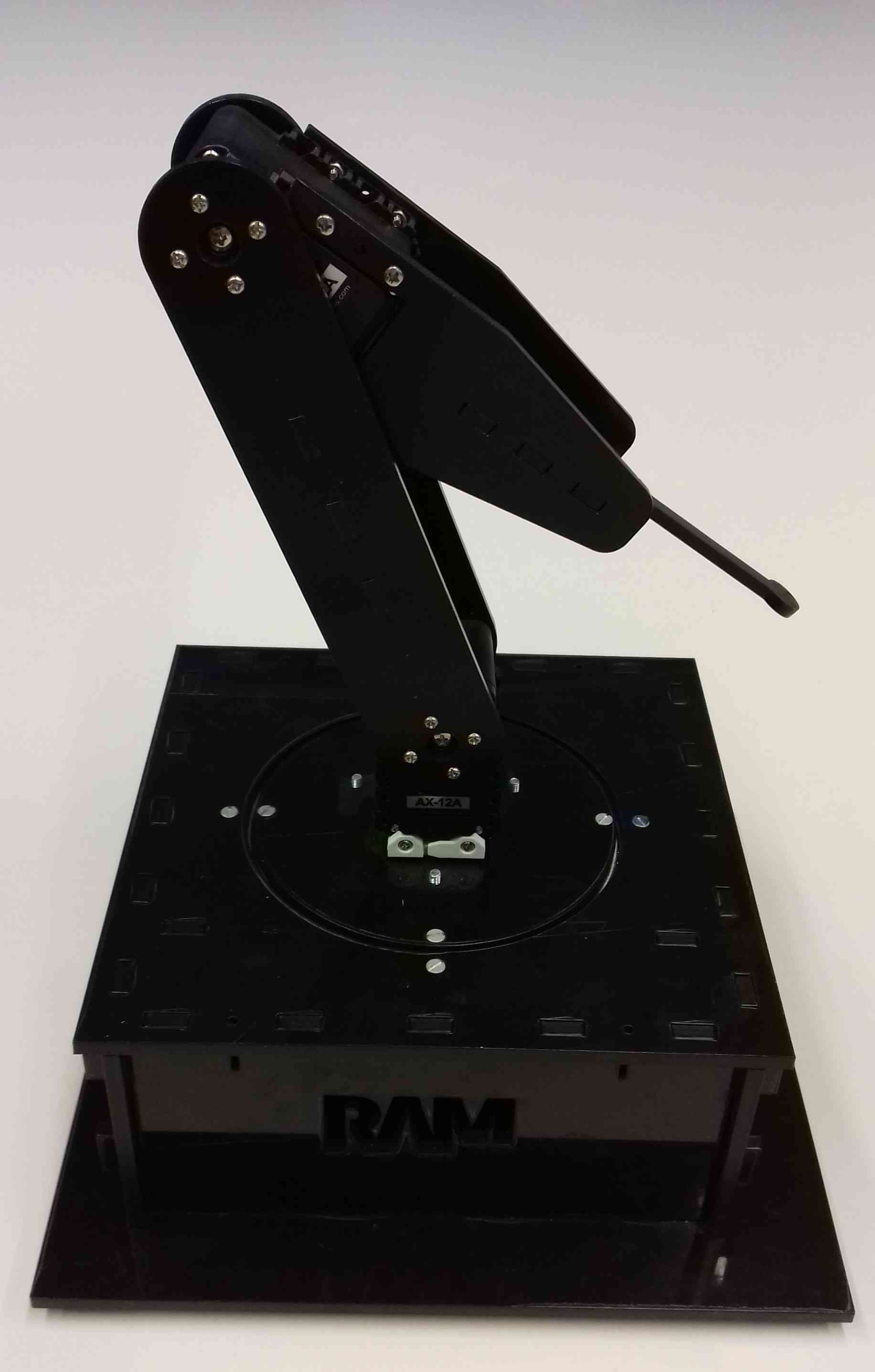}} 
\hfill
\subfloat[][Pusher task.]{\includegraphics[height=0.28\linewidth]{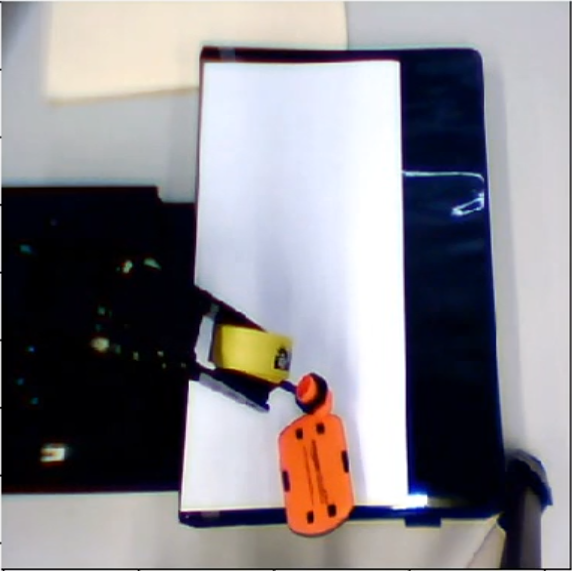}} 
\hfill
\subfloat[][Reacher task.]{\includegraphics[height=0.28\linewidth]{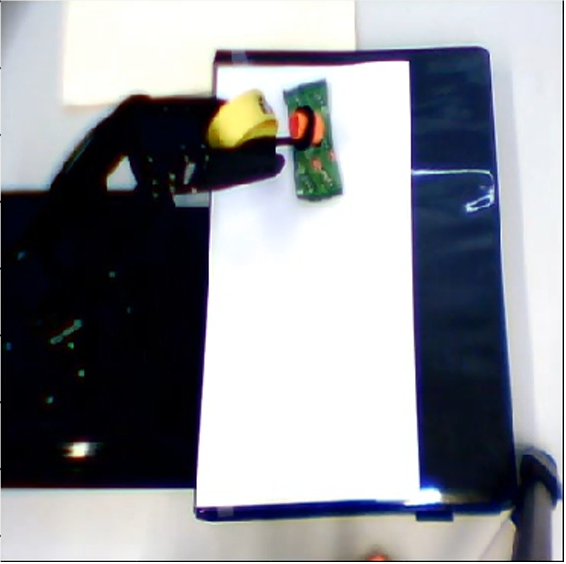}}
\hfill\null
\caption{Pusher/Reacher.}
\label{fig:PusherReacher}
\end{figure}

\textbf{Pusher/Reacher:} Two validation tasks with a 3DoF robotic arm (see Fig.~\ref{fig:PusherReacher}). The problems of pushing  and reaching an object were addressed.  
For both tasks the robot arm is placed in front of the work-space and an RGB camera is fixed overhead for capturing the top-down view of the environment with images of $640\times480\times3$ size. The images are downsampled to $64\times64\times3$. The objective of the Pusher task is to move the object placed in the work-space down, until it is out, as depicted in Fig.~\ref{fig:PusherReacher}(b). The objective of the Reacher is to track the position of the object with the arm's end effector (Fig.~\ref{fig:PusherReacher}(c)). In these problems, the teacher advises corrections of the position commands of the arm in the Cartesian space. The experiments of the tasks with the 3DoF robot arm were intended only to validate the proposed learning method in another real setup, no comparisons were carried out.

All the results that present averaged data in the form of a curve have confidence intervals that represent the $60^{th}$ percentile of the data.
The neural network hyperparameters proposed in \cite{perez2018interactive} were used in this work. The experiments are illustrated in the attached video\footnote{https://youtu.be/i4f1D4CH26E}. 

\subsection{Study with simulated teacher}
In order to evaluate the method along with subtle variants under more controlled conditions, a high performance policy standing-in as a teacher, which was actually trained with D-COACH and a real human teacher, was used (similar approach to the one used in  \cite{Celemin2018AnInteractive}). The simulated teacher generates feedback computing $h = \operatorname{sign}(a_\mathrm{teacher} - a_\mathrm{agent})$, whereas the decision on whether to provide feedback at each time step is given by the probability $P_{h} = \alpha \cdot\exp(-\tau\cdot \mathrm{timestep})$, where $\{\alpha \in {\rm I\!R}\ | 0 \le \alpha \le 1\}$; $\{\tau \in {\rm I\!R}\ | 0 \le \tau\}$.

To perform an ablation study that evaluates the contribution of the new components of the enhanced D-COACH, the complete method (Algorithm \ref{algorithm:DeepCOACH}) is compared to a second variant that does not include the AE contribution, and learns the whole policy network only with the cost of predicting the action. This variant  works like the basic D-COACH proposed in \cite{perez2018interactive}, but skipping the first two steps of recording demonstrations and pre-training the AE, i.e., learning from scratch, which means that the convolutional layers are also learned with the teacher's corrections.

The third evaluated variant includes the AE cost function, but never freezes the convolutional layers, so it always modifies the parameters of the complete policy network using the gradient of both cost functions (i.e. setting $\epsilon=0$ for the condition in line 6). The learning curves of the three cases of D-COACH are compared against the baseline one of a DDPG-based RL agent \cite{Lillicrap2015}, using the OpenAI implementation \cite{baselines}. The curves are the average of 30 runs for each case, showing the evolution of the return through the learning time. The time considered is measured when rendering the environments, i.e., no environment acceleration, since D-COACH is intended for learning with real systems wherein speeding up the environment is not possible.

\begin{figure}[t]
\centering
\subfloat[][Car Racing.]{\includegraphics[width=0.5\linewidth]{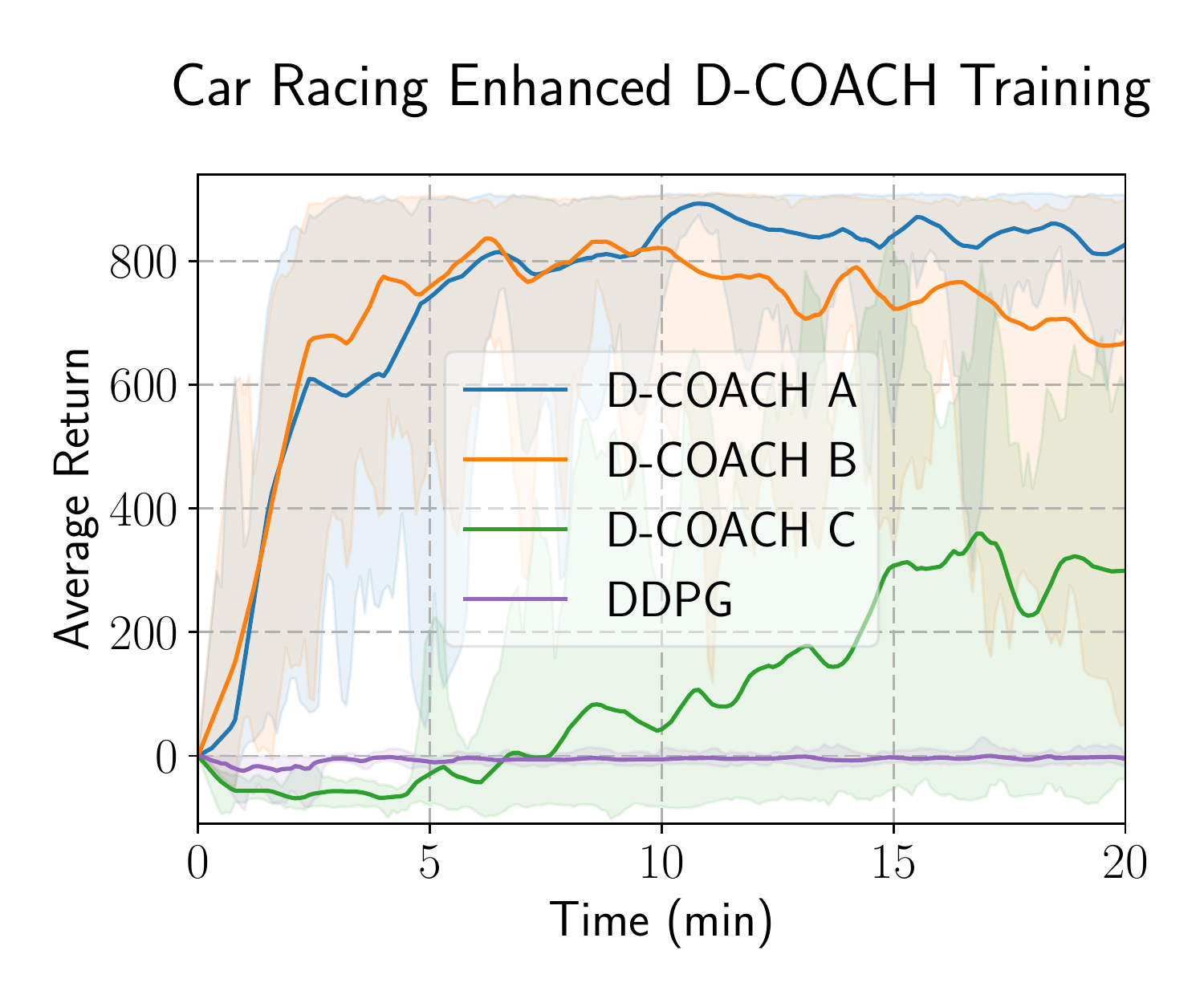}}
\subfloat[][Duckie Racing.]{\includegraphics[width=0.5\linewidth]{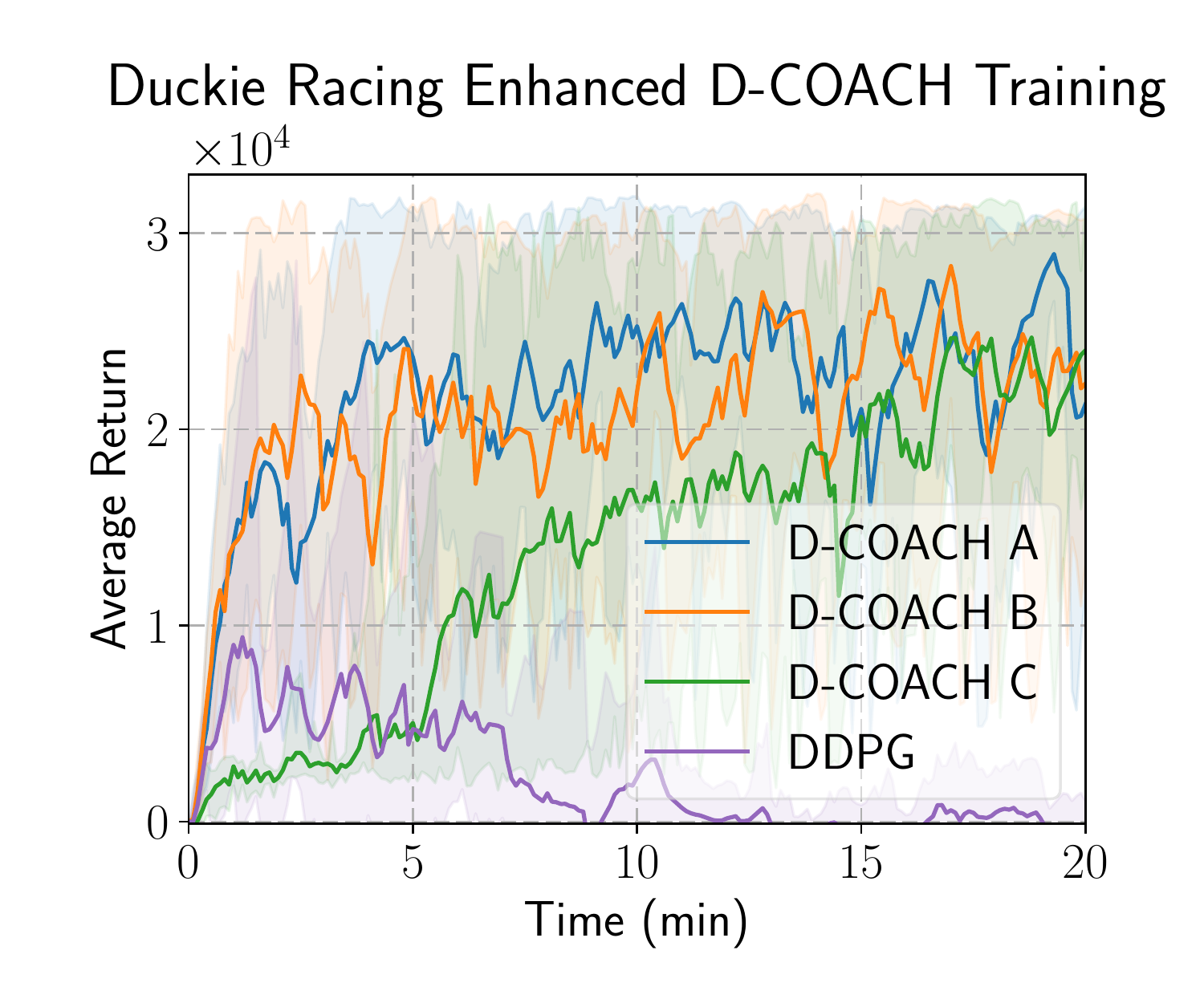}}
\caption{Car/Duckie Racing results for simulated teacher with D-COACH and DDPG. D-COACH A: policy and AE costs, freezing conv. layers; D-COACH B: policy and AE costs; D-COACH C: only policy cost. Buffer: $K = 1000$; $k=20$; $b = 10$; $N = 8$. $P_{h}$: $\alpha = 0.6$; $\tau = 0.000015$.} 
\label{fig:simulatedteachers} 
\end{figure}

As shown in the Fig.~\ref{fig:simulatedteachers} for the experiments with the Car Racing, the complete D-COACH has a considerable improvement when simultaneously using the gradients of the auto-encoding cost function for learning the state representation, along with the gradient of the policy  (blue and orange curves), in contrast to using only the gradient of the policy (green curve), which is slower, and reaches less than $50\%$ of outcome with respect to the complete algorithm after 20 minutes of training. Additionally, there is no noticeable improvement in the performance for the RL agent within this time frame.

The results of the experiments with the Duckie Racing problem (Fig.~\ref{fig:simulatedteachers}) show similar trends as observed with the Car Racing problem, wherein the contribution of the AE cost function makes a considerable difference with respect to only using the policy cost. However, in this problem the variant of D-COACH using only the policy cost manages to reach the same level of performance of the other variants after 17 minutes of training. This variant can learn good policies for this problem, but for reaching $95\%$ of the final performance, it is around 5 times slower than the variants using simultaneous auto-encoding. For this problem again the DDPG learning process does not obtain any improvement during the first 20 minutes of learning process. 

Finally, it is possible to see the contribution of the condition stated for freezing the convolutional layers, when the error of the decoder is small. This rule provides more stability to the learning process. In the Car Racing experiments, the variant that always updates the AE undergoes an ``unlearning'' stage after 10 minutes of training, whereas in the Duckie Racing experiments is not possible to notice any considerable difference between both approaches. When the error of the decoder is small it means that the latent vector is a good representation of the state, but still the gradient and the error of the policy can be large; therefore, in some cases there may be conflicts that harm the AE performance and consequently the performance of the policy. Freezing these layers is a detail that solves this conflict.

\subsection{Experiments with real human teachers}

The experiments with simulated teachers are useful for analyzing the evolution of the learning process. However, D\nobreakdash-COACH is an interactive learning method; therefore, it is necessary to carry out experiments with real human teachers for complementing its evaluation. Specifically, we perform experiments for measuring the human effort in terms of the time dedicated to teach the agent. The experiments compare the basic D-COACH and the enhanced D-COACH, evaluating the necessary effort (time) to achieve some levels of performance. Ten participants between 20 and 27 years old were asked to act as teachers for both the Car Racing and the Duckie Racing problem. In each problem, the participants corrected the agent's actions with the arrow symbols of a keyboard for a limited session of 20 minutes. 
The average results are presented and discussed.

\begin{figure}[t]
\centering
\subfloat[][Car Racing.]{\includegraphics[width=0.5\linewidth]{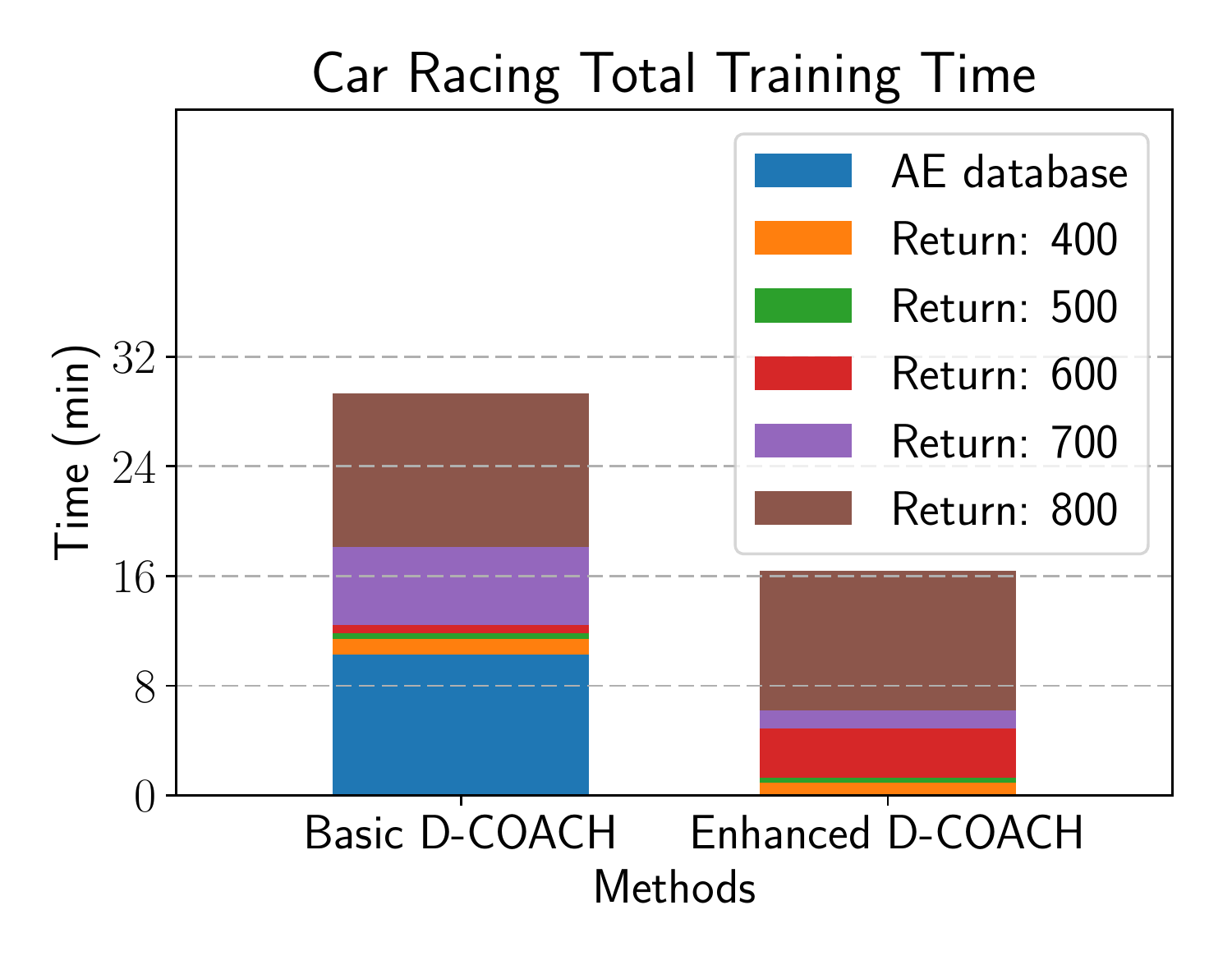}}
\subfloat[][Duckie Racing.]{\includegraphics[width=0.5\linewidth]{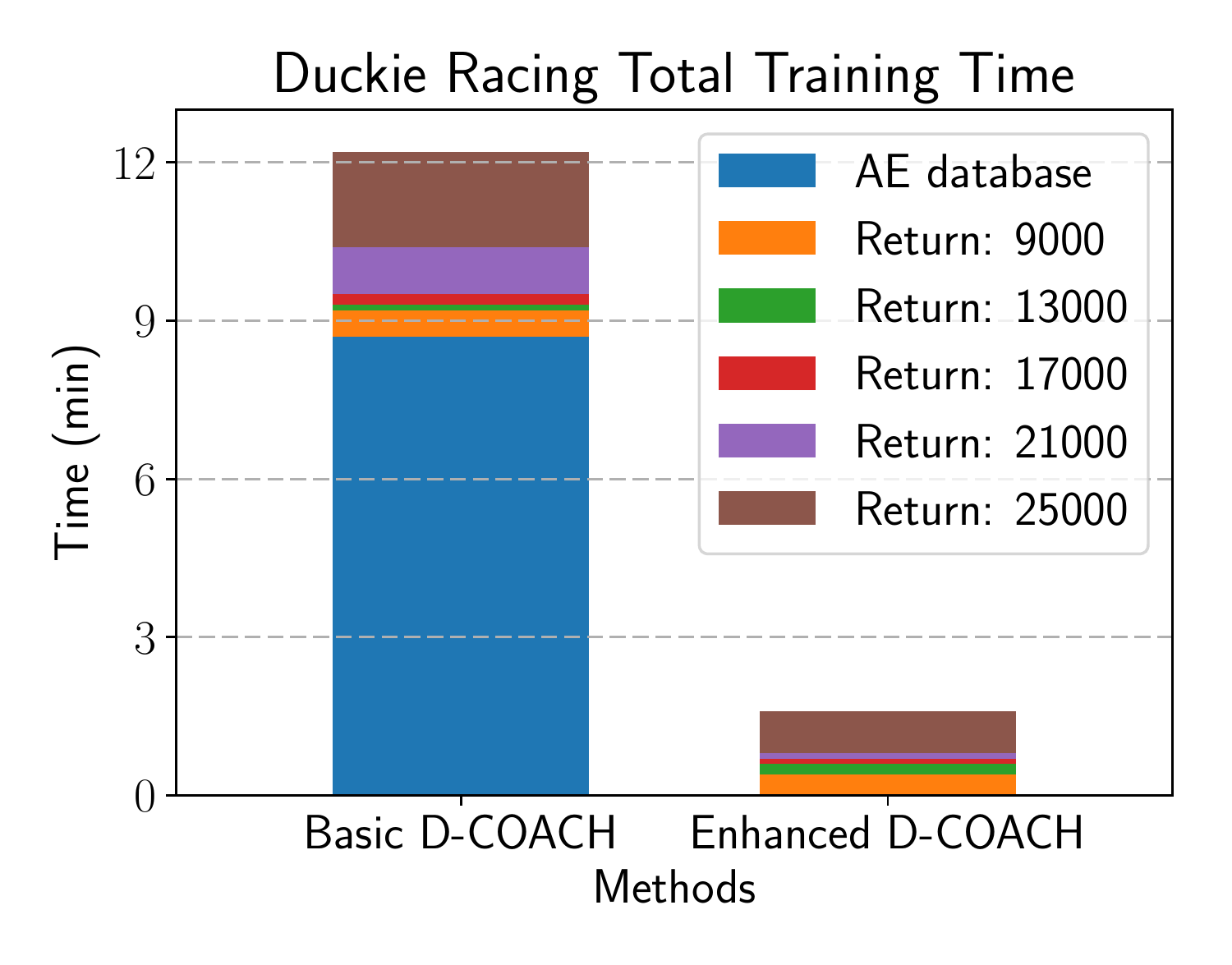}}
\caption{Comparison of the average human time dedicated to achieve for the first time some levels of return (on average).} 
\label{fig:stacked_bar} 
\end{figure}

\begin{figure}[h]
\centering
\subfloat[][Car Racing.]{\includegraphics[width=0.5\linewidth]{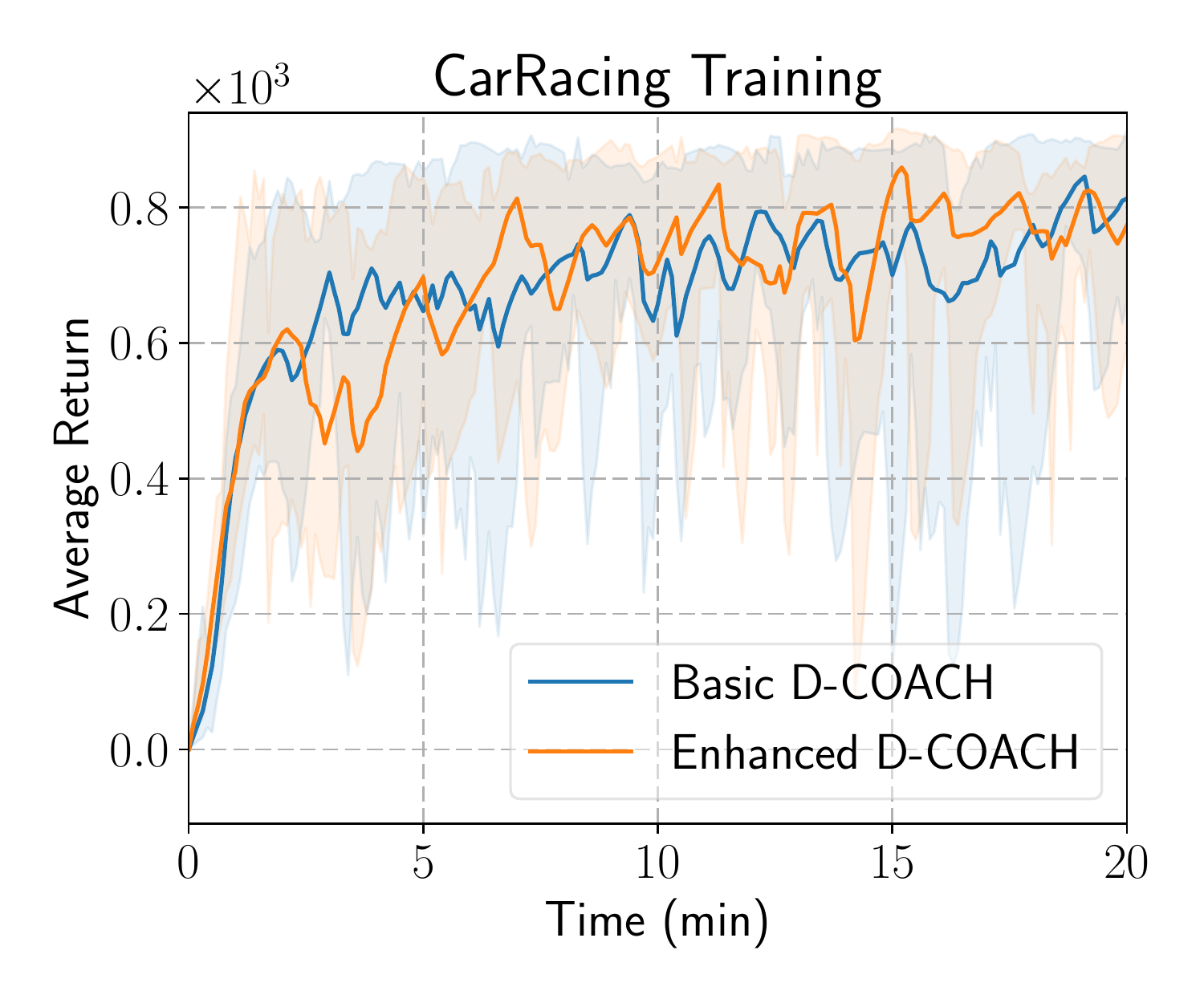}}
\subfloat[][Duckie Racing.]{\includegraphics[width=0.5\linewidth]{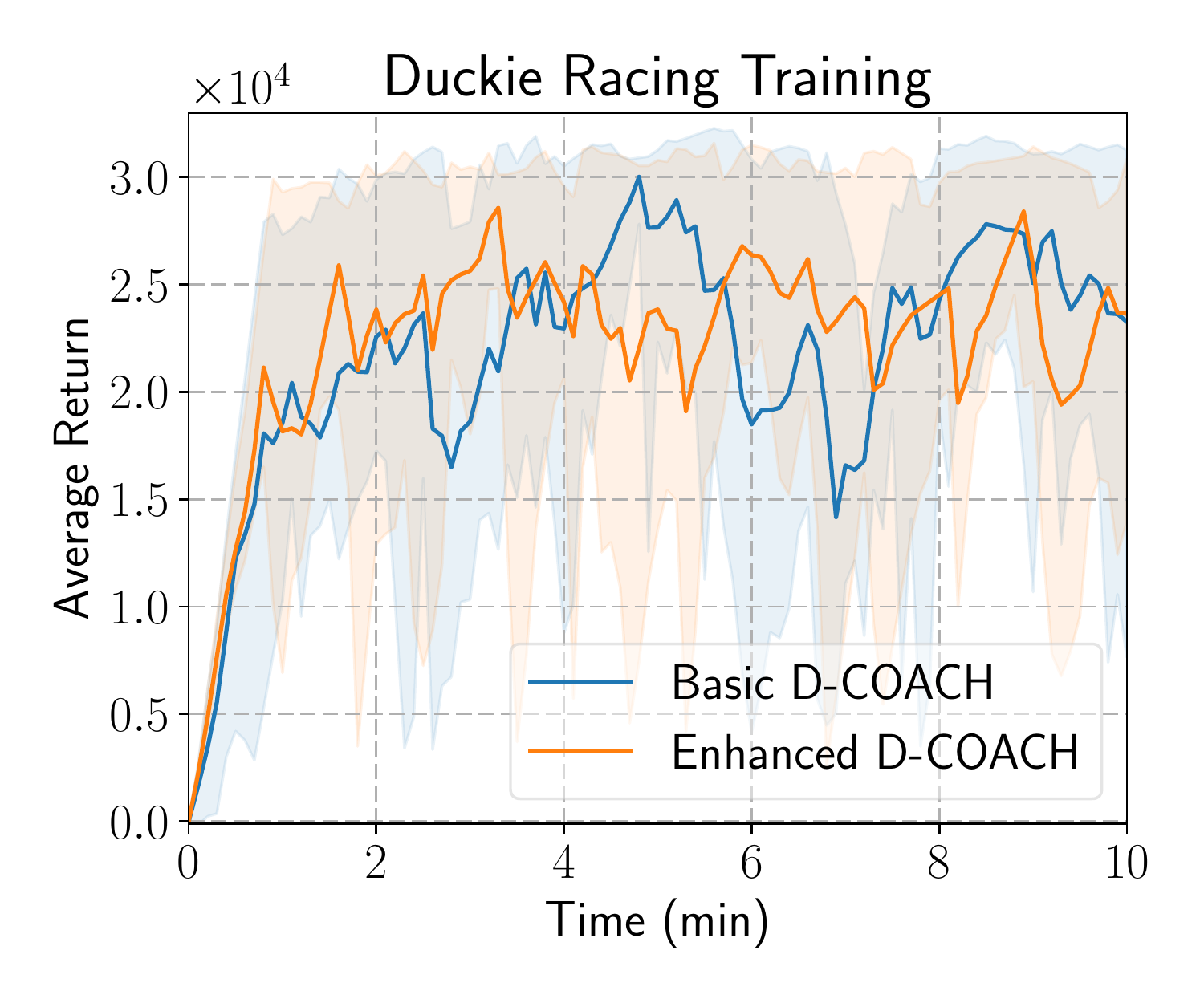}}
\caption{Results of learning with human teachers. Buffer: $K = 1000$; $k=20$; $b = 10$; $N = 8$.} 
\label{fig:humanteachers} 
\end{figure}

In Fig.~\ref{fig:stacked_bar} the time dedicated for training the agents is depicted. In the cases of learning with the basic D-COACH, the blue bar indicates the time dedicated by the teachers in its first step of recording demonstrations, which for both problems is actually longer than the time used for reaching the highest level of performance with the enhanced D-COACH. In total, the new method saves around $45\%$ of the training time for the Car Racing problem, and above $80\%$ for the Duckie Racing problem. These results do not include the time dedicated to train the AE in the basic D\nobreakdash-COACH, which would depend on the available hardware. The bar diagram is complemented with the learning curves in Fig.~\ref{fig:humanteachers} (for the basic D-COACH the curve is only after training the AE), wherein it is shown that the enhanced D-COACH has a similar progress with a very slight advantage over its basic version, even without considering the additional time required for the AE training step.

\subsection{Additional validation with real systems}

Additional experiments with human teachers interacting with real robots through the enhanced D-COACH were carried out. These tests are for validating the results obtained with the previous two types of experiments, and no comparisons are presented.

A real Duckiebot was used for validating the results obtained with the simulations. An experienced teacher advised the policy of Duckiebot from scratch and obtained a good policy in six minutes.

Similarly, in the pusher/reacher problems well performing policies were obtained within twenty minutes. 

\begin{figure}[t]
    \centering
    \includegraphics[width=\linewidth]{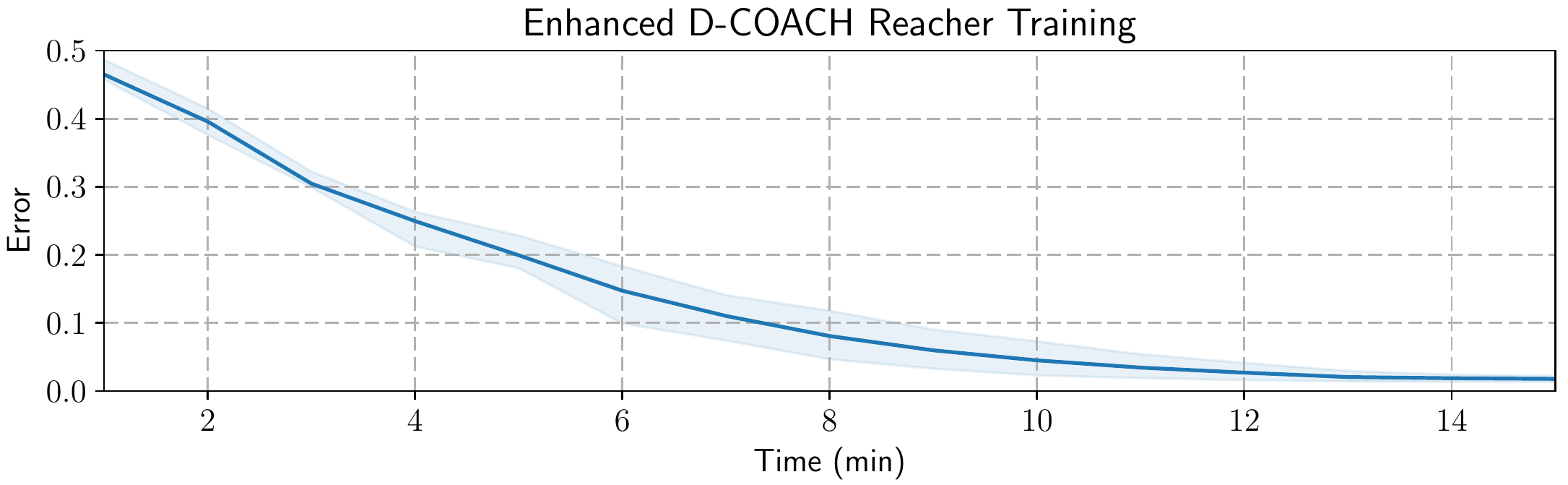}
    \caption{Evolution of the error while learning the reacher task. }
    \label{fig:reacher_exp}
\end{figure}

Fig.~\ref{fig:reacher_exp} shows an extra validation that was done for the reacher case. A cost function was defined as the Euclidean distance between the end-effector of the arm and the object to track, normalized with the largest possible distance within the image (distance of opposite corners). Seven training sessions of 15 minutes were run and averaged. It is possible to observe that the cost decreases as the learning process advances. 

\section{Conclusions}
\label{sec:conclusion}

This work has introduced an improved version of an interactive method for training policies represented with deep neural networks, particularly for problems wherein the observed state is defined in a high-dimensional space like a raw image.

The proposed enhanced D-COACH offers a simpler learning scheme of only one step in which state representation and the policy itself are learned jointly using the two optimization criteria (the AE cost and the regression error of the policy). This method eliminates the necessity of recording demonstrations for the pre-training of the AE, which is a time consuming effort for the user, and sometimes is not possible due to complexity of the problem and lack of complex skills of the user in the task domain. In the approached problems, the effort of the users was reduced between $45\%$ and $80\%$. The enhanced D-COACH can adapt and extract features to represent reached unknown states during the learning process, which would be problematic for its basic version. Additionally, computational effort is reduced with the possibility of skipping the offline training of the AE, which is usually expensive. 

This simultaneous method is very data efficient for training the state representation. The AE is trained with the data gathered when human teachers advise corrections, which ensures a very representative database. Those samples correspond to the most important regions of the state space wherein the policy needs to discriminate different actions to execute.

The results also show that the interactive method can obtain higher performances than DRL in very few episodes. The level of the performance achieved by the interactive method would be obtained by DRL agents after several hundreds or thousands of episodes, which means that our proposed method is actually feasible, for learning with real robots in many applications wherein RL is not yet.

\bibliographystyle{IEEEtran}
\bibliography{biblio}

\begin{thebibliography}{10}
\providecommand{\url}[1]{#1}
\csname url@samestyle\endcsname
\providecommand{\newblock}{\relax}
\providecommand{\bibinfo}[2]{#2}
\providecommand{\BIBentrySTDinterwordspacing}{\spaceskip=0pt\relax}
\providecommand{\BIBentryALTinterwordstretchfactor}{4}
\providecommand{\BIBentryALTinterwordspacing}{\spaceskip=\fontdimen2\font plus
\BIBentryALTinterwordstretchfactor\fontdimen3\font minus
  \fontdimen4\font\relax}
\providecommand{\BIBforeignlanguage}[2]{{%
\expandafter\ifx\csname l@#1\endcsname\relax
\typeout{** WARNING: IEEEtran.bst: No hyphenation pattern has been}%
\typeout{** loaded for the language `#1'. Using the pattern for}%
\typeout{** the default language instead.}%
\else
\language=\csname l@#1\endcsname
\fi
#2}}
\providecommand{\BIBdecl}{\relax}
\BIBdecl

\bibitem{atari}
\BIBentryALTinterwordspacing
V.~Mnih, K.~Kavukcuoglu, D.~Silver, A.~Graves, I.~Antonoglou, D.~Wierstra, and
  M.~Riedmiller, ``Playing atari with deep reinforcement learning,'' 2013, cite
  arxiv:1312.5602Comment: NIPS Deep Learning Workshop 2013. [Online].
  Available: \url{http://arxiv.org/abs/1312.5602}
\BIBentrySTDinterwordspacing

\bibitem{Silver2016}
D.~Silver, A.~Huang, C.~J. Maddison, A.~Guez, L.~Sifre, G.~van~den Driessche,
  J.~Schrittwieser, I.~Antonoglou, V.~Panneershelvam, M.~Lanctot, S.~Dieleman,
  D.~Grewe, J.~Nham, N.~Kalchbrenner, I.~Sutskever, T.~Lillicrap, M.~Leach,
  K.~Kavukcuoglu, T.~Graepel, and D.~Hassabis, ``Mastering the game of go with
  deep neural networks and tree search,'' \emph{nature}, vol. 529, no. 7587, p.
  484, 2016.

\bibitem{brockman2016openai}
G.~Brockman, V.~Cheung, L.~Pettersson, J.~Schneider, J.~Schulman, J.~Tang, and
  W.~Zaremba, ``Openai gym,'' \emph{arXiv preprint arXiv:1606.01540}, 2016.

\bibitem{tassa2018deepmind}
Y.~Tassa, Y.~Doron, A.~Muldal, T.~Erez, Y.~Li, D.~d.~L. Casas, D.~Budden,
  A.~Abdolmaleki, J.~Merel, A.~Lefrancq \emph{et~al.}, ``Deepmind control
  suite,'' \emph{arXiv preprint arXiv:1801.00690}, 2018.

\bibitem{Gu2017}
S.~Gu, E.~Holly, T.~Lillicrap, and S.~Levine, ``Deep reinforcement learning for
  robotic manipulation with asynchronous off-policy updates,'' in \emph{IEEE
  International Conference on Robotics and Automation (ICRA), 2017}.\hskip 1em
  plus 0.5em minus 0.4em\relax IEEE, 2017, pp. 3389--3396.

\bibitem{koos2013transferability}
S.~Koos, J.-B. Mouret, and S.~Doncieux, ``The transferability approach:
  Crossing the reality gap in evolutionary robotics,'' \emph{IEEE Transactions
  on Evolutionary Computation}, vol.~17, no.~1, pp. 122--145, 2013.

\bibitem{akrour2011preference}
R.~Akrour, M.~Schoenauer, and M.~Sebag, ``Preference-based policy learning,''
  in \emph{Joint European Conference on Machine Learning and Knowledge
  Discovery in Databases}.\hskip 1em plus 0.5em minus 0.4em\relax Springer,
  2011, pp. 12--27.

\bibitem{Knox:2009:ISA:1597735.1597738}
\BIBentryALTinterwordspacing
W.~B. Knox and P.~Stone, ``Interactively shaping agents via human
  reinforcement: The tamer framework,'' in \emph{Proceedings of the Fifth
  International Conference on Knowledge Capture}, ser. K-CAP '09.\hskip 1em
  plus 0.5em minus 0.4em\relax New York, NY, USA: ACM, 2009, pp. 9--16.
  [Online]. Available: \url{http://doi.acm.org/10.1145/1597735.1597738}
\BIBentrySTDinterwordspacing

\bibitem{Celemin2018AnInteractive}
C.~Celemin and J.~Ruiz-del Solar, ``An interactive framework for learning
  continuous actions policies based on corrective feedback,'' \emph{Journal of
  Intelligent \& Robotic Systems}, pp. 1--21, 2018.

\bibitem{perez2018interactive}
R.~Perez, C.~Celemin, J.~Ruiz-del Solar, and J.~Kober, ``Interactive learning
  with corrective feedback for policies based on deep neural networks,'' in
  \emph{International Symposium on Experimental Robotics}.\hskip 1em plus 0.5em
  minus 0.4em\relax Springer, 2018.

\bibitem{Chernova2014}
S.~Chernova and A.~L. Thomaz, \emph{Robot learning from human teachers}.\hskip
  1em plus 0.5em minus 0.4em\relax Morgan \& Claypool Publishers, 2014, vol.~8,
  no.~3.

\bibitem{Argall2009}
B.~D. Argall, S.~Chernova, M.~Veloso, and B.~Browning, ``A survey of robot
  learning from demonstration,'' \emph{Robotics and autonomous systems},
  vol.~57, no.~5, pp. 469--483, 2009.

\bibitem{macglashan2017interactive}
J.~MacGlashan, M.~K. Ho, R.~Loftin, B.~Peng, G.~Wang, D.~L. Roberts, M.~E.
  Taylor, and M.~L. Littman, ``Interactive learning from policy-dependent human
  feedback,'' in \emph{Proceedings of the 34th International Conference on
  Machine Learning-Volume 70}.\hskip 1em plus 0.5em minus 0.4em\relax JMLR.
  org, 2017, pp. 2285--2294.

\bibitem{Christiano2017}
P.~F. Christiano, J.~Leike, T.~Brown, M.~Martic, S.~Legg, and D.~Amodei, ``Deep
  reinforcement learning from human preferences,'' in \emph{Advances in Neural
  Information Processing Systems}, 2017, pp. 4299--4307.

\bibitem{Warnell2017}
\BIBentryALTinterwordspacing
G.~Warnell, N.~Waytowich, V.~Lawhern, and P.~Stone, ``{Deep TAMER: Interactive
  Agent Shaping in High-Dimensional State Spaces},'' sep 2017. [Online].
  Available: \url{http://arxiv.org/abs/1709.10163}
\BIBentrySTDinterwordspacing

\bibitem{Argall2008}
B.~D. Argall, B.~Browning, and M.~Veloso, ``Learning robot motion control with
  demonstration and advice-operators,'' in \emph{2008 IEEE/RSJ International
  Conference on Intelligent Robots and Systems}.\hskip 1em plus 0.5em minus
  0.4em\relax IEEE, 2008, pp. 399--404.

\bibitem{Mnih2015}
V.~Mnih, K.~Kavukcuoglu, D.~Silver, A.~A. Rusu, J.~Veness, M.~G. Bellemare,
  A.~Graves, M.~Riedmiller, A.~K. Fidjeland, G.~Ostrovski \emph{et~al.},
  ``Human-level control through deep reinforcement learning,'' \emph{Nature},
  vol. 518, no. 7540, p. 529, 2015.

\bibitem{zhang2017deep}
T.~Zhang, Z.~McCarthy, O.~Jow, D.~Lee, K.~Goldberg, and P.~Abbeel, ``Deep
  imitation learning for complex manipulation tasks from virtual reality
  teleoperation,'' \emph{arXiv preprint arXiv:1710.04615}, 2017.

\bibitem{ross2011reduction}
S.~Ross, G.~Gordon, and D.~Bagnell, ``A reduction of imitation learning and
  structured prediction to no-regret online learning,'' in \emph{Proceedings of
  the fourteenth international conference on artificial intelligence and
  statistics}, 2011, pp. 627--635.

\bibitem{Finn2015}
C.~Finn, X.~Y. Tan, Y.~Duan, T.~Darrell, S.~Levine, and P.~Abbeel, ``Deep
  spatial autoencoders for visuomotor learning,'' in \emph{2016 IEEE
  International Conference on Robotics and Automation (ICRA)}.\hskip 1em plus
  0.5em minus 0.4em\relax IEEE, 2016, pp. 512--519.

\bibitem{Ha2018}
D.~Ha and J.~Schmidhuber, ``World models,'' \emph{arXiv preprint
  arXiv:1803.10122}, 2018.

\bibitem{Paull2017}
L.~Paull, J.~Tani, H.~Ahn, J.~Alonso-Mora, L.~Carlone, M.~Cap, Y.~F. Chen,
  C.~Choi, J.~Dusek, Y.~Fang, D.~Hoehener, S.~Y. Liu, M.~Novitzky, I.~F.
  Okuyama, J.~Pazis, G.~Rosman, V.~Varricchio, H.~C. Wang, D.~Yershov, H.~Zhao,
  M.~Benjamin, C.~Carr, M.~Zuber, S.~Karaman, E.~Frazzoli, D.~{Del Vecchio},
  D.~Rus, J.~How, J.~Leonard, and A.~Censi, ``{Duckietown: An open, inexpensive
  and flexible platform for autonomy education and research},'' in
  \emph{Proceedings - IEEE International Conference on Robotics and
  Automation}, 2017, pp. 1497--1504.

\bibitem{gym_duckietown}
M.~Chevalier-Boisvert, F.~Golemo, Y.~Cao, B.~Mehta, and L.~Paull, ``Duckietown
  environments for openai gym,''
  \url{https://github.com/duckietown/gym-duckietown}, 2018.

\bibitem{Lillicrap2015}
T.~P. Lillicrap, J.~J. Hunt, A.~Pritzel, N.~Heess, T.~Erez, Y.~Tassa,
  D.~Silver, and D.~Wierstra, ``Continuous control with deep reinforcement
  learning,'' \emph{arXiv preprint arXiv:1509.02971}, 2015.

\bibitem{baselines}
P.~Dhariwal, C.~Hesse, O.~Klimov, A.~Nichol, M.~Plappert, A.~Radford,
  J.~Schulman, S.~Sidor, and Y.~Wu, ``Openai baselines,''
  \url{https://github.com/openai/baselines}, 2017.

\end{thebibliography}

\end{document}